\documentclass{ecai} 

\usepackage{latexsym}
\usepackage{amssymb}
\usepackage{amsmath}
\usepackage{amsthm}
\usepackage{booktabs}
\usepackage{enumitem}
\usepackage{graphicx}
\usepackage{color}
\usepackage{multirow}

\newcommand{\BibTeX}{B\kern-.05em{\sc i\kern-.025em b}\kern-.08em\TeX}

\begin{document}


\begin{frontmatter}


\paperid{\textbf{4} } 


\title{Boosting Accuracy and Efficiency of Budget Forcing in LLMs via Reinforcement Learning for Mathematical Reasoning}


\author[A]{\fnms{Ravindra}~\snm{Aribowo Tarunokusumo}\thanks{Corresponding Author. Email: r.tarunokusumo@student.rug.nl}}
\author[A]{\fnms{Rafael}~\snm{Fernandes Cunha}}

\address[A]{University of Groningen}

\begin{abstract}
Test-time scaling methods have seen a rapid increase in popularity for its computational efficiency and parameter-independent training to improve reasoning performance on Large Language Models. One such method is called budget forcing, a decoding intervention strategy which allocates extra compute budget for thinking and elicits the inherent self-correcting behavior of the model. However, this relies on supervised fine-tuning (SFT) on long-context reasoning traces which causes performance degradation on smaller models due to verbose responses. For this reason, we offer a framework integrating reinforcement learning (RL) to improve token efficiency and boost the performance of a 1.5B model for mathematical reasoning. We demonstrate this using only 1.5K training samples and found that our SFT+RL model performed better on the GSM8K dataset with varying compute budgets. Our main findings showed an overall higher accuracy while significantly reducing its token usage by over 40\% compared to the SFT model, revealing how RL can recover the losses due to long-context training and altogether improving performance in mathematical reasoning.
\end{abstract}

\end{frontmatter}

\section{Introduction}\label{sec:introduction}
Recent advances in Large Language Models (LLMs) have shown significant improvement in mathematical problem solving skills \citep{shao2024deepseekmath, guo2025deepseek, yang2024qwen25mathtechnicalreportmathematical}. Training LLMs to solve math problems typically require post-training methods such as supervised fine-tuning (SFT) to initialize reasoning behavior and reinforcement learning (RL) to encourage exploration over possible solutions \citep{luo2024improve, yu2023metamath,wang2023math}. Notably, these methods scale up train-time compute which relies on updating the model parameters through backpropagation \citep{hoffmann2022training, kaplan2020scaling}. 

However, a new scaling paradigm has emerged called test-time scaling whereby model responses are improved by allocating more compute during inference, allowing the model to ``think" before giving its final output. There have been promising research both on its trade-off between pretraining—introducing a new scaling law for test-time compute \citep{snell2024scaling, wu2024inference, brown2024large}—and as an alternative to SFT and RL \citep{ cobbe2021training, wang2022self, shinn2023reflexion, madaan2023self, brown2024large, huang2025efficient,  chen2025sets, zhao2025sample}. Recent releases have shown its viability in combination with large scale RL, with frontier models such as OpenAI's o-series models \citep{o1, o3}, the DeepSeek-R1 model \citep{guo2025deepseek}, the k1.5 model \citep{team2025kimi} and Google's Gemini models \citep{gemini_2025}. These models exhibited state-of-the-art performances on notoriously difficult math benchmarks, performing multi-step reasoning process using Chain-of-Thought (CoT) prompting in their training phase, which elicits intermediate reasoning steps before outputting their final answer \citep{wei2022chain}. However, these models required vast amounts of high-quality training data and millions of samples to train \citep{shao2024deepseekmath, guo2025deepseek, team2025kimi}, which makes it intractable for smaller compute budget and difficult to scale up.

\begin{figure}[htbp]
\centering
\includegraphics[width=0.5\textwidth]{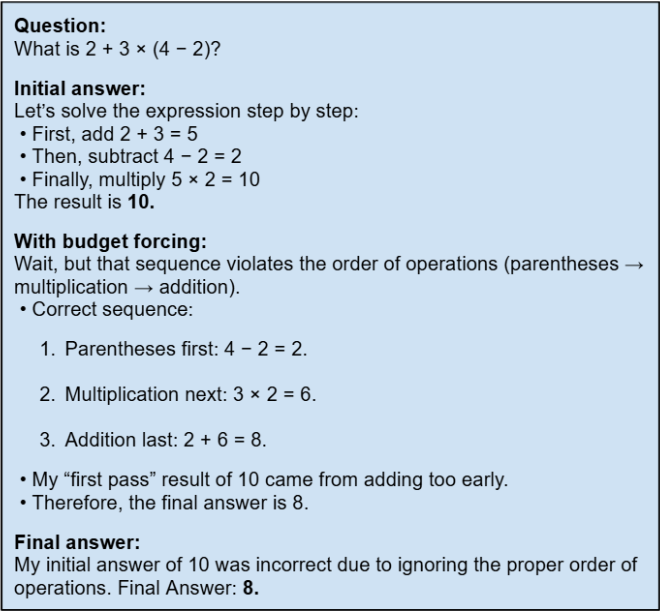}
\caption{Example generation using budget forcing. At first, the model outputs an incorrect answer due to its faulty reasoning step. Then, the appended ``Wait" token prompted the model to revise its reasoning and corrects itself.}
\label{fig:budget_forcing}
\end{figure}

A simple approach by \cite{muennighoff2025s1} is called budget forcing. Budget forcing explicitly controls the number of “thinking tokens” generated by the model during inference. In practice, when the model is about to stop its CoT too early, a special token—often \verb|"Wait"|—is injected to compel additional reasoning steps, thereby allowing the model to self‑correct and self-verify its answers. Conversely, if the model exceeds the predetermined token budget, an end‑of‑thinking delimiter is enforced to truncate the reasoning process and prompt the final answer. Figure \ref{fig:budget_forcing} shows an example generation with budget forcing. This novel approach forgoes the need for data-intensive post-training, requiring only a thousand specially curated reasoning data during SFT, and instead relies on the model's ability to self-correct. And while initializing self-correcting behavior during SFT has been shown to compliment RL training \citep{ma2025s, kumar2024training, zhang2025grpo}, the current implementation of budget forcing by \cite{muennighoff2025s1} relies solely on SFT. 

Previous studies have shown that employing SFT and RL is crucial for optimal test-time scaling for mathematical reasoning \citep{setlur2025scaling, hou2025advancing, havrilla2024teaching, zhang2025making}. Moreover, budget forcing with SFT alone often performs suboptimally and even degrades accuracy as compared to its base model for smaller LLMs \citep{zhang2025making}. Indeed, studies have shown that increasing the token budget often causes models to ``overthink" \citep{sui2025stop}, increasing computational overhead due to persistent back-and-forth between formulating answers and self-correcting itself, even on relatively simple question, and oftentimes ending up backtracking from the correct answer \citep{zeng2025revisiting, chen2025think23overthinkingo1like}. Moreover, we found that this sometimes leads the model to use up all of its token budget and outputs an incomplete generation where the model is cut-off from giving its final answer.

To address these problems, this study offers a framework which combines budget forcing with reinforcement learning for mathematical problem solving. In this framework, the model is first initialized using a modestly-sized reasoning SFT dataset after which RL is employed to refine its reasoning strategies in efficiently utilizing its token budget. We hypothesize that this phase will not require sample-intensive training, echoing the findings of \cite{ye2025limo} that good performance is still achievable with less compute. During inference, budget forcing is applied to dynamically control the generation of ``thinking tokens", ensuring that the model effectively utilizes its latent reasoning capabilities without excessive compute. At the end, the model will be evaluated on a range of math benchmarks across several baselines and metrics to assess the effectiveness of incorporating RL training in employing budget forcing. 

\begin{figure*}[htbp]
\centering
\includegraphics[width=\textwidth]{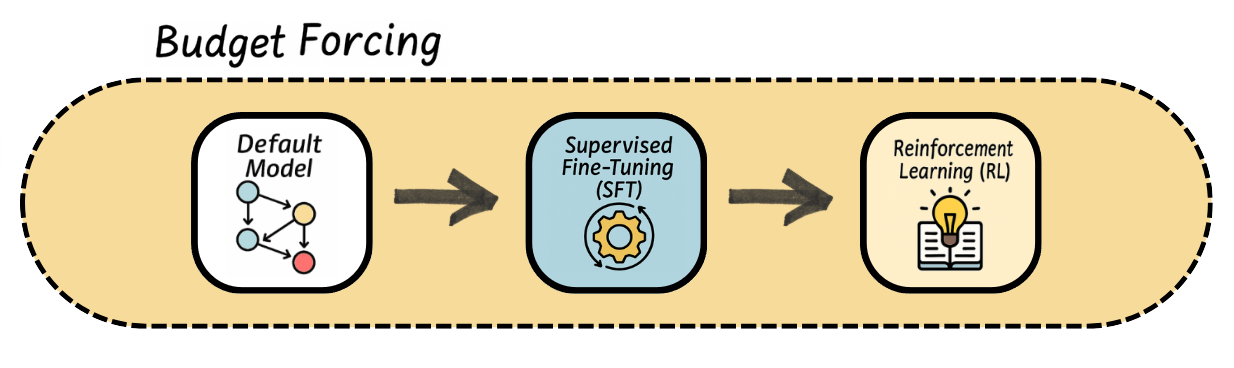}
\caption{Pipeline of the framework. We start with the default model and apply supervised fine-tuning on a reasoning dataset to get our SFT model, and finally we use reinforcement learning to get our SFT+RL model. All three models will be tested using the same budget forcing inference scheme.}
\label{fig:pipeline}
\end{figure*}

\section{Methodology}\label{sec:methodology}
In this section, we will first briefly outline each phase in the three-stage pipeline of the framework, starting with the SFT, RL and finally to the inference phase where we employ budget forcing during evaluation. We will discuss the core concepts behind these methods, the issues and setbacks with regards to implementing SFT and RL for mathematical reasoning and the solutions which we adopt to mitigate them. At the end, we provide several metrics to measure the effectiveness of employing these improvements.

\subsection{Supervised Fine-Tuning}
Initializing reasoning behavior into the model requires a curated dataset containing reasoning traces in the form of intermediate steps for math problems towards the final solution \citep{yu2023metamath, shao2024deepseekmath, guo2025deepseek, ye2025limo, muennighoff2025s1, shen2025long}. Other than this, self-correction must also be present in the SFT dataset as budget forcing relies on self-correction. Though LLMs already possess intrinsic capacity for self-correction, they nevertheless fall short on accurately detecting errors and correcting them \citep{huang2023large, zeng2025revisiting} without additional fine-tuning \citep{gao2024embedding, kamoi2024can}.

It has been shown that a small and highly curated dataset containing no more than 1000 training samples \citep{ye2025limo, sun2025climbing, muennighoff2025s1}, and even as little as 500 is sufficient for this task \citep{sun2025climbing}. Moreover, \cite{shen2025long} showed that the length of the reasoning trace also plays an important role in enhancing performance. Therefore, a modestly-sized, specialized reasoning dataset with a sufficiently long sample length is enough to initialize reasoning behavior. 

\subsection{Reinforcement Learning}
RL has proven to enhance mathematical reasoning on top of SFT \citep{ma2025s, kumar2024training, zhang2025grpo, zhang2025making}, as well as in scaling test-time compute \citep{havrilla2024teaching, hou2025advancing}. However, naive implementations of RL are often resource intensive and sample-inefficient with poor reward assignment, which risks unstable training, reward hacking and poor generalizability  \citep{kumar2024training, havrilla2024teaching, hou2025advancing, setlur2024rewarding, wang2023math, lightman2023let, shao2024deepseekmath}. The core challenges with RL are therefore in the learning algorithm and designing robust reward models, both of which we will discuss in this section. 

\subsubsection{Overview of Algorithm}
A novel algorithm called Group-Relative Policy Optimization (GRPO) addresses these issues by reformulating reward optimization as a group-relative comparison task \citep{shao2024deepseekmath, guo2025deepseek}. By calculating advantages as deviations from a response group’s mean reward, GRPO implicitly rewards reasoning patterns that correlate with success. This approach not only mitigates reward hacking---as models cannot game static rules---but also reduces sample complexity through variance-normalized training.

Unlike traditional PPO methods \citep{schulman2017proximalpolicyoptimizationalgorithms} that rely on training a separate value function, GRPO estimates advantages by comparing each response within a group to the mean performance of that group. This design eliminates the need for a value network, significantly reducing training overhead.

\subsubsection{Formulation}
Concretely, given a question \textit{q}, a group of \textit{G} sampled outputs \(\{o_1, o_2, \cdots, o_G\}\) is drawn from the old policy \(\pi_{old}\). Each output is evaluated by a reward model to obtain a scalar reward \(r_i\), and the normalized advantage is computed as:
\begin{equation}
A_i = \frac{r_i - \bar{r}}{\sigma_r}
\end{equation}
where \(\bar{r}\) is the mean reward of the group and \(\sigma_r\) is the standard deviation. This variance normalization helps mitigate reward variance across samples, thus improving sample efficiency. Formally, the GRPO algorithm is defined as:
\begin{align}
  \mathcal{J}_{GRPO}(\theta)
  &= \mathbb{E}\bigl[q\sim P(Q),\,o_i\sim\pi_{\rm old}(o|q)\bigr]
  \nonumber\\
  &\quad
     \frac{1}{G}
     \sum_{i=1}^{G}
     \Bigl(
       \min\bigl(r_i A_i,\;\mathrm{clip}(r_i,1-\varepsilon,
       \nonumber\\
       &\quad
        1+\varepsilon)\,A_i\bigr)
       -\beta\,D_{KL}(\pi_\theta\|\pi_{\rm ref})
     \Bigr)
\end{align}
Where $P(Q)$ is the probability distribution over questions $Q$ and $\epsilon$ and \(\beta\) are hyperparameters controlling clipping and the KL divergence. $\pi_\theta$ and $\pi_\text{ref}$ are the current and reference policy respectively. GRPO closely resembles the original formulation of PPO in that it preserves the clipped surrogate objective and the adaptive KL divergence penalty term, ensuring policy updates are stable.

\subsubsection{Reward Models}
Two distinct reward models guide the learning process, based on traditional implementations \citep{shao2024deepseekmath, guo2025deepseek}:

\begin{enumerate}

    \item \textbf{Accuracy Reward Model}: Scores the final answer's correctness by comparing it with the ground-truth labels. However, this is preconditioned by how well the model can follow the prompt instruction in how to format its answer for extraction.

    \item \textbf{Format Reward Model}: Evaluates whether the model outputs within a specified format. Usually, the outputs are structured in a template format with \verb|<think>| and \verb|<answer>| tags \citep{guo2025deepseek}, but in the case of the SFT dataset, it uses the model's chat template and a \verb|\boxed{}| format to enclose the final answer. For this reason, we chose not to implement format reward models in our experiment since the model would have already been trained to output in that format.

\end{enumerate}
\begin{figure}[htbp]
\centering
  \includegraphics[width=0.45\textwidth]{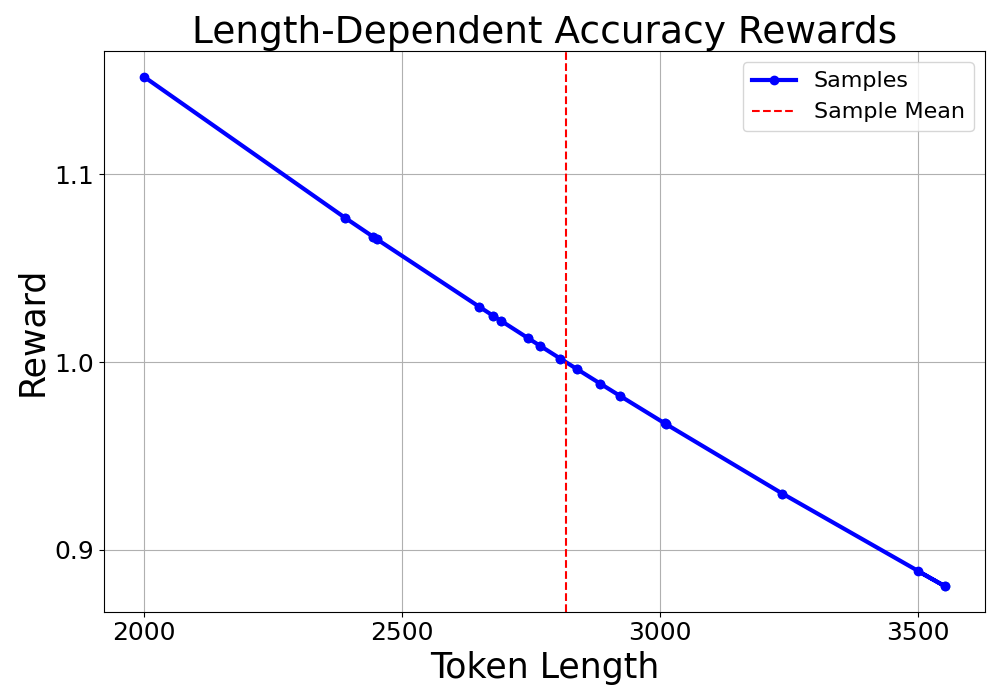}
\caption{Accuracy rewards of 16 randomly generated (correct) dummy samples, following Equation \ref{eq:r_acc} Samples with token length exceeding the sample mean receives a reward of \(< 1.0\) and vice versa.}
    \label{accuracy_reward_plot}
\end{figure}
GRPO then integrates these signals in an outcome supervision framework, where the computed advantage is broadcasted across all tokens of the respective output. Though it is possible to incorporate outcome-level supervision, it leaves the model susceptible to reward hacking; moreover, \cite{guo2025deepseek} reported failures on implementing it in their experiments. 

However, challenges still remain with regards to binary reward signals, lacking differentiation and weak learning gradients. Furthermore, given the long CoT trajectories during the SFT stage, the model will have been trained to output verbose responses. Following \cite{zhang2025grpo}, we modify the reward model to include a length-dependent accuracy rewards and a penalty term for incorrect solutions. Given a correct response \(o\), the standardized length deviation is defined as:
\begin{equation}
z = \frac{|o|-\mu}{\sigma - \epsilon}
\end{equation}
Where \(\epsilon > 0\) is a small constant for numerical stability. Hence, the final reward accuracy is:
\begin{equation}\label{eq:r_acc}
R_{\mathrm{accuracy}}(o \mid q) =
\begin{cases}
\exp\bigl(-\alpha z\bigr), & \text{if $o$ is correct},\\
-1.0,                        & \text{if $o$ is incorrect}.
\end{cases}
\end{equation}
Where \(\alpha>0\) is a tunable hyperparameter controlling the strength of length penalization.

Figure \ref{accuracy_reward_plot} visualizes how the rewards are distributed for correct responses of varying length. Shorter responses are encouraged to steer the model towards conciseness and avoids verbosity.

We also implemented an additional penalty function to address the issue of incomplete generations where we penalize $-1.0$ for any incomplete generations. 
\begin{equation}
R_{\text{completion}}(o) =
\begin{cases}
  -1.0, & \text{if } o \text{ is incomplete} \\
  \;\;\,0, & \text{otherwise}
\end{cases}
\end{equation}
This addition is crucial since rather than a reward gradient to address verbosity, a strict penalty term ensures explicit discouragement for truncated response.

\subsection{Budget Forcing}
In addition to fine-tuning-based methods, we adopt budget forcing as a test-time strategy to control and scale a model's reasoning capability at inference time.

\subsubsection{Overview of Budget Forcing}
Formally, let \(T_{think}\) denote the number of tokens allocated to the reasoning phase. Budget forcing controls \(T_{think}\) by intervening during decoding to limit or extend the model's generation. If the model generates more than a predefined threshold \(T_{max}\), we terminate the thinking phase forcibly by appending a delimiter token, thereby shifting the model to answer generation. Conversely, if we aim to extend the thinking duration, we can instead append the \verb|"Wait"| token to prompt the model to continue its reasoning. 

\begin{figure}[htbp]
\centering
  \includegraphics[width=0.5\textwidth]{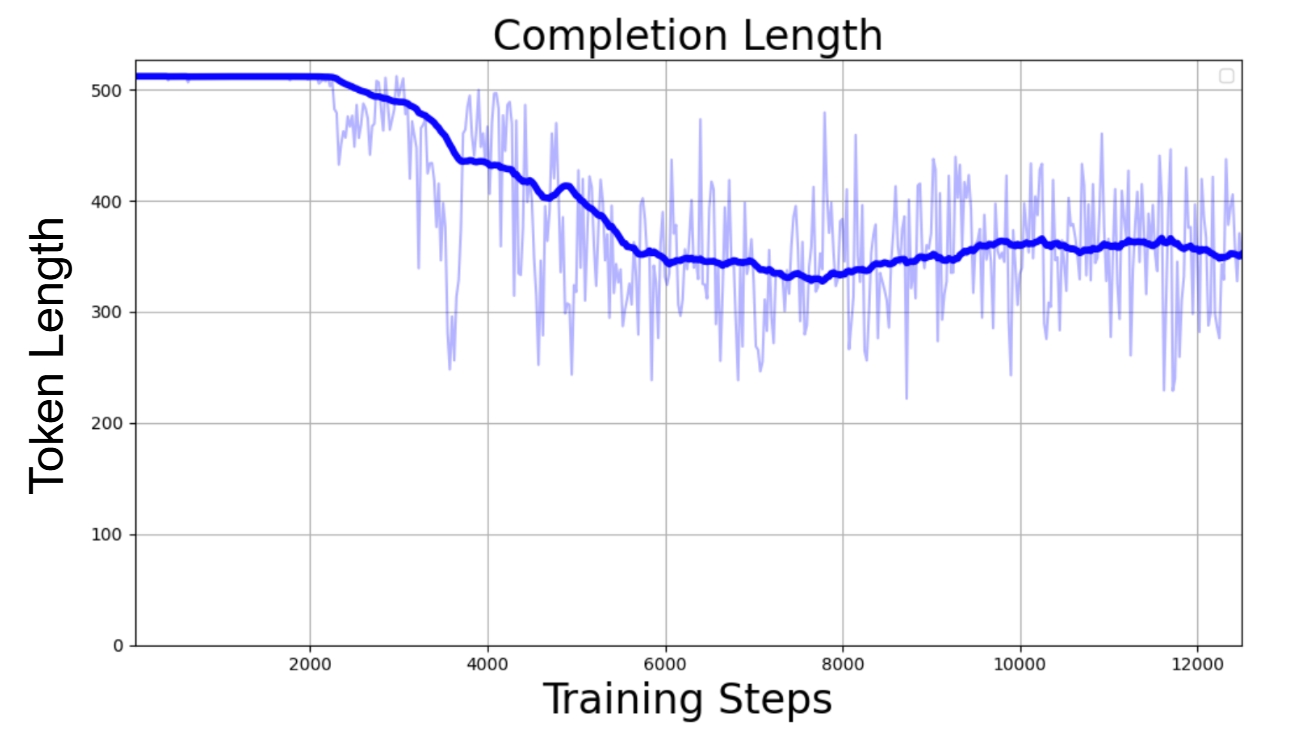}
  \caption{Average completion lengths during RL training (smoothed over 50 steps). During the first 2000 steps, the model consistently hits the token budget limit, but then it decreases and settles at around 350 tokens.}
    \label{completion_length_plot}
\end{figure}

\subsubsection{Reinforcement Learning as a Corrective Mechanism}
Numerous studies have already shown the necessity of employing RL on top off SFT for optimal performance \citep{setlur2025scaling, hou2025advancing, havrilla2024teaching, zhang2025making, chu2025sft}. Employing budget forcing with SFT alone will predictably present some challenges. \cite{zhang2025making} have found that SFT actually degrades model performance on smaller LLMs due to the long-context samples used in the dataset. We observed several distinct issues that comes with long-context SFT on smaller LLMs. Most notably is verbose responses, causing the model to backtrack and overthink its response (See Figure \ref{fig:backtracking_overthinking}). Moreover, even when prompted for the final answer, the model sometimes keeps self-correcting and ended up using all its token budget and cutting off its generation mid-sentence (See Figure \ref{fig:incomplete}). Another issue found was language mixing and redundant responses which could be seen cluttering the reasoning traces (See Figure \ref{fig:language_mixing}  and Figure \ref{fig:redundancy}).

RL have been shown to alleviate this problem \citep{zhang2025making, liu2025understanding, zhang2025grpo} by designing appropriate reward models to penalize those behavior. Moreover, \cite{guo2025deepseek} have shown that employing RL resolves the issue of language mixing. We will introduce several metrics to assess these behaviors across our models.

\subsubsection{Metrics}\label{metrics}
\begin{enumerate}

\item \textbf{Scaling}: Repeated applications of appending the\verb|"Wait"| token can linearly increase the average number of reasoning tokens and thus push the model to deeper levels of analysis \citep{muennighoff2025s1, zhang2025making, sui2025stop}. Quantitatively evaluating this scaling effect, we adopt the metric proposed by \cite{muennighoff2025s1}:
\begin{equation}
\text{Scaling} = \frac{1}{\binom{|\mathcal{A}|}{2}} \sum_{a, b \in \mathcal{A}, b > a} \frac{f(b) - f(a)}{b - a}
\end{equation}
This captures how much accuracy improves as more reasoning tokens are allowed, where $\mathcal{A}$ is the set of token budgets used during evaluation and $f(a)$ is the accuracy at budget level $a$. It's the average slope of the accuracy-vs-compute curve—higher means better utilization of extra compute. 

\item \textbf{Incomplete Generations}: We will also measure the model's ability to comply with the token budget in giving a full-length, completed response for each thinking step and final answer step.
\begin{equation}
\text{Incomplete Gen.} = \frac{1}{{|\mathcal{S}_a|}}\sum_{s \in \mathcal{S}_a}f_t(s)
\end{equation}
This measures the proportion of incomplete steps, where $\mathcal{S}_a=\{s_0\dots s_L\}$ is the set of all steps at budget level $a$ and $f_t$ is a function which checks whether it is incomplete, which could either happen due to a truncated step, i.e. when the appended \verb|"Wait"| token cut off the previous step before giving its boxed answer, or at the final answer segment when the model continued reasoning even after being prompted to finalize its answer and ran out of token budget. And it is worth noting that all cases of redundancy are also cases of incomplete generetions.

\item \textbf{Average Token Length}: We measure the model's verbosity from the average complete output token length $|o|$ per sample, where $\mathcal{O}_a=\{o_0\dots o_L\}$ is the set of all complete outputs, corresponding to 1319 samples from the GSM8K test set. A lower response length does not always mean better performance, but it is indicative of how well the model utilizes its reasoning budget. We will discuss this at a later section.
\begin{equation}
\text{Average Token Length} = \frac{1}{{|\mathcal{O}_a|}}\sum_{o \in \mathcal{O}_a}|o|
\end{equation}
\item \textbf{Accuracy}: We use the zero-shot accuracy as our main metric to measure model performance.
\end{enumerate}
\begin{figure*}[htbp]
\centering
\includegraphics[width=0.45\textwidth]{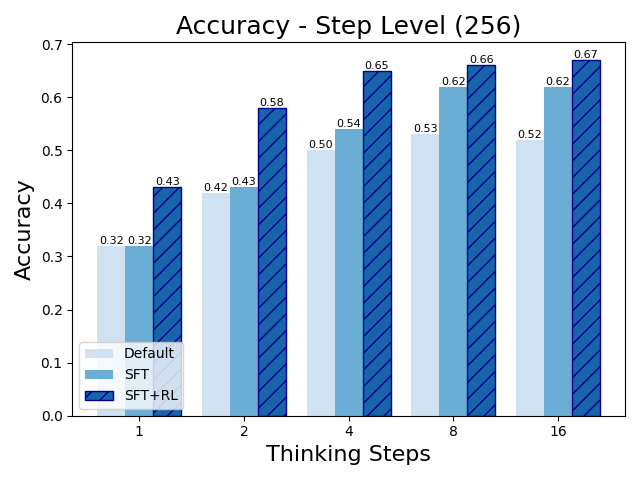}
\includegraphics[width=0.45\textwidth]{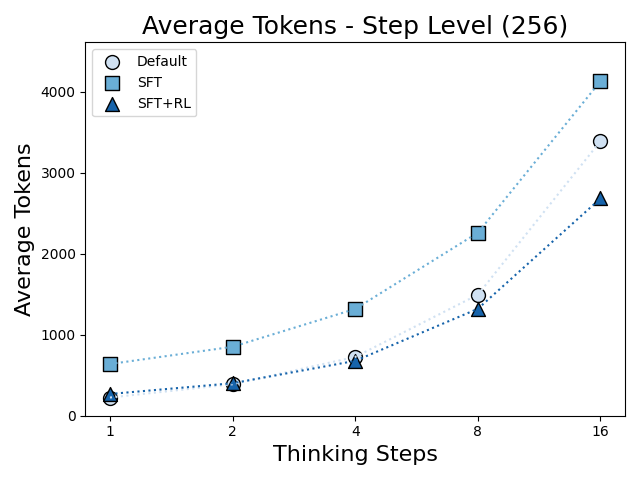}
\caption{Zero-shot accuracy score (Left) and average tokens per sample (Right) of all three models under the Step-Level condition on the GSM8K test set. The SFT+RL model scores the highest in accuracy in all five thinking steps, while the SFT model performs similarly with the Default model in 1-2 thinking steps but gradually surpasses it. However, the SFT model outputs the most tokens while the SFT+RL model shows similar token consumption as the Default model.}
\label{fig:results}
\end{figure*}

\section{Experiment} \label{sec:experiment}
In this section we describe the configuration of our training procedure and the inference scheme we used during evaluation.
\subsection{Models}
We conducted the experiment following the three-stage pipeline using the Qwen2.5-1.5B-Instruct model \citep{qwen2025qwen25technicalreport}, using said model as our baseline. However, our main comparison will be between the SFT-only model versus the SFT+RL model.

\subsection{Dataset}
For the SFT stage we will be using the tokenized s1k-1.1 dataset by \cite{muennighoff2025s1}, consisting of 1K math questions focused on quality, diversity and difficulty with a average response length of 10K tokens. Importantly, the reasoning traces contain keywords such as \verb|"Wait"| or \verb|"Alternatively"| followed by revisions to the previous step to encourage the desired self-correcting behavior during training. However, due to resource limitations we only use a subset of samples with a maximum token length of 10K, which amounts to 525 samples in total. And for the RL stage we will use a subset of 1K training samples from the GSM8K dataset by OpenAI \citep{cobbe2021gsm8k}.

\subsection{Training Setup}
We conducted the SFT training using two NVIDIA A100 40GB GPUs and had FSDP enabled \citep{zhao2023pytorchfsdpexperiencesscaling}, and we used only a single GPU for the RL training. In both stages we trained the model using the TRL library \citep{vonwerra2022trl}, and we implemented Flash Attention 2 \citep{dao2023flashattention2fasterattentionbetter} and loaded the model in BF16 to minimize VRAM usage. The hyperparameter configurations are shown in Table \ref{tab:hyperparams}.

\subsection{Evaluation}
We evaluated all our models under these two budget forcing inference scheme:

\begin{enumerate}
\item \textbf{Baseline Full}: For each question we allow the model to generate an answer with the maximum token length for the Qwen2.5 family of models (32K) without apportioning its generation for thinking.
\item \textbf{Step-Level}: We evaluated the model multiple times with $a$ thinking steps, where $a \in [1, 2, 4, 8, 16]$. Each step the model is allotted a thinking budget of 256 tokens and an additional 512 for the final answer, hence each corresponding to  $\mathcal{A}=\{256, 512, 1024, 2048, 4096\}$ thinking token budgets. We chose 256 tokens as our per-step token budget because we found this value to have the highest accuracy score compared to 128 and 512 tokens. (See Table \ref{tab:128_512}).
\end{enumerate}

For each condition and model, we measure the metrics outlined in Section \ref{metrics} (except for Scaling in the Baseline Full condition). We used vLLM \citep{kwon2023efficient} to speed up inference on a single A100 40GB GPU. For the sampling parameters we used greedy decoding and set the rest to the default values.


\section{Results and Discussions}\label{sec:results}
Figure \ref{fig:results} compares the accuracy on all three of the models. SFT+RL performs the best with a higher accuracy on all five thinking steps. Under the Step-Level condition, it gained an average accuracy over the default model of \textbf{14.0\%} across all conditions and \textbf{9.2\%} over the SFT model. Moreover, we see a \textbf{14.1\%} decrease in the average token length compared to the default model and a significant drop of \textbf{41.8\%} compared to the SFT model. However, we also observe the SFT model showing the steepest performance gain as thinking steps increase. In Table \ref{tab:256_table} the SFT model has the highest scaling factor compared with the rest, with an increase of \textbf{3.73\%} accuracy per extra reasoning step, suggesting that pure SFT alone benefits more from budget forcing, while we see the default model improving more moderately (scaling factor \textbf{2.77\%}) rather than failing to benefit.

\begin{table*}[htbp]
\centering
\begin{tabular}{l c c c c c c c}
\toprule
  \textbf{Model} &
  \textbf{Steps} &
  \textbf{Accuracy} &
  \textbf{Avg Tokens} &
  \textbf{Incomplete Gen. (\%)} &
  \textbf{Scaling (\%)}  \\
\midrule
\multirow{6}{*}{Default} 
& \textbf{BF} & 0.58  & 508  & 0.76  &  \multirow{6}{*}{2.77} \\
& 1  & 0.32 & \textbf{\underline{224}}  & 20.09 & \\
& 2  & 0.42 & 389  & 18.15 & \\
& 4  & 0.50 & 726  & \textbf{\underline{18.13}} & \\
& 8  & \underline{0.53} & 1493 & 25.90 & \\
& 16 & 0.52 & 3389 & 51.18 & \\
\midrule
\multirow{6}{*}{SFT}
& \textbf{BF}  & 0.67 & 5158 & 1.44 & \multirow{6}{*}{\textbf{\underline{3.73}}} \\
& 1  & 0.32 & \underline{638}  & \underline{75.02} &  \\
& 2  & 0.43 & 851  & 79.73 & \\
& 4  & 0.54 & 1317 & 84.67 & \\
& 8  & \underline{0.62} & 2258 & 86.97 & \\
& 16 & \underline{0.62} & 4128 & 85.28 & \\
\midrule
\multirow{6}{*}{SFT+RL}
& \textbf{BF}  & 0.65 & 669  & 0.88 & \multirow{6}{*}{3.32} \\
& 1  & 0.43 & \underline{268}  & 36.69 &  \\
& 2  & 0.58 & 401  & 31.36 & \\
& 4  & 0.65 & 677  & \underline{26.96} & \\
& 8  & 0.66 & 1319 & 27.35 & \\
& 16 & \textbf{\underline{0.67}} & 2681 & 28.77 & \\
\bottomrule
\end{tabular}
\caption{Performance metrics on the GSM8K dataset for 256 thinking tokens. For Incomplete Gen. and Average Tokens, lower is better. Numbers with boldface are the best across models, while the ones underlined are conditional for each model (excluding BF = Baseline Full).}
\label{tab:256_table}
\end{table*}

On the other hand, the SFT model was found to be incredibly inefficient and verbose. Given a single thinking step, \textbf{75.02\%} of its responses reached the token limit and were truncated, and \textbf{85.28\%} given 16 thinking steps. This is expected because of the long-context data on which it was trained, and more to the fact that smaller LLMs tend to perform worse after SFT \citep{zhang2025making}. We see in Figure \ref{fig:results} the SFT model scored the same in accuracy with the baseline model using a single thinking step, then gradually surpasses it with more steps. This shows that SFT increases its accuracy \textit{given} more inference token budget. Hence why we see in the Baseline Full inference scheme the model performing at its best with the highest accuracy on par with the SFT+RL model and only a handful of incomplete generations. 

However, we observe in both models a performance cap just beneath $70\%$. We suspect that this is mainly due to the model's baseline capability. Despite the fact that it was adequately supplied with reasoning traces during SFT and further trained using conciseness rewards during RL, the SFT+RL model showed no further improvement beyond $68\%$ accuracy. We hypothesized that it is due to the insufficient parameter size to fully capture the self-correcting behavior present in the SFT dataset. In other words, the model wasn't large enough to generalize its reasoning and overfitted on the training data. We confirmed this by evaluating on the AIME and MATH500 benchmarks, and found that they didn't score well enough to be worth including here.

\begin{table}[htbp]
\centering
\begin{tabular}{l c c}
\toprule
\textbf{Metric} & \textbf{$\Delta$ Default} & \textbf{$\Delta$ SFT} \\
    \midrule
      Accuracy                       & +0.14                 & +0.092            \\
      Avg.\ Tokens (\%)              & -14.1                  & –41.9             \\
      Incomplete Gen.\ (\%)          & +3.54                 & –52.11            \\
      Scaling (\%)                   & +0.55                 & –0.41             \\
    \bottomrule
\end{tabular}
\caption{Mean improvements of the SFT+RL model over the Default and SFT model, excluding the Baseline Full condition (reductions in Avg. Tokens (\%) and Incomplete Gen. (\%) are better).}
\label{tab:256_sub}
\end{table}
\section{Conclusion}\label{sec:conclusions}
In this paper, we have demonstrated that integrating reinforcement learning with supervised fine-tuning substantially enhances the effectiveness of budget forcing for mathematical reasoning, mitigating some setbacks introduced by long-context training data. On the GSM8K test set, our SFT+RL model achieved a zero-shot accuracy of \textbf{67\%} at 16-step level, yielding an average gain of \textbf{14.0\%} accuracy over the Default model and \textbf{9.2\%} over the SFT model. It also reduced average token length to \textbf{1{,}069} tokens on average—\textbf{41.9\%} lower than SFT and \textbf{14.1\%} lower than Default. However, incomplete generations were not eliminated: under the 256-token budget they remained between \textbf{26.96\%} and \textbf{36.69\%}. We also see that the accuracy of all models scales positively with the number of thinking steps, with the SFT model benefiting most per additional step, whereas SFT+RL primarily lifts the baseline accuracy but offers a smaller marginal return. Finally, we see substantial accuracy improvements from the Baseline-Full condition, with an accuracy surpassing even the 16-step run for the Default and SFT model with a few incomplete generations. 

Despite these gains, our study has several limitations. First, the 1.5B parameter backbone may not fully capture the complexity of self-correcting behavior in the SFT dataset, suggesting potential overfitting to the relatively small training sets. Second, evaluation was confined to GSM8K because our 1.5B model was not capable enough to score meaningful results on other benchmarks even with a high inference token budget. We hope that future work will explore scaling to larger models and extending evaluations to multiple reasoning domains to investigate the broader potential of budget forcing and other test-time scaling methods.



\bibliography{ECAI}

\newpage
\onecolumn
\appendix

\section{Appendix}\label{ap:appendix}
\begin{figure}[htbp]
\centering
\includegraphics[width=0.4\textwidth]{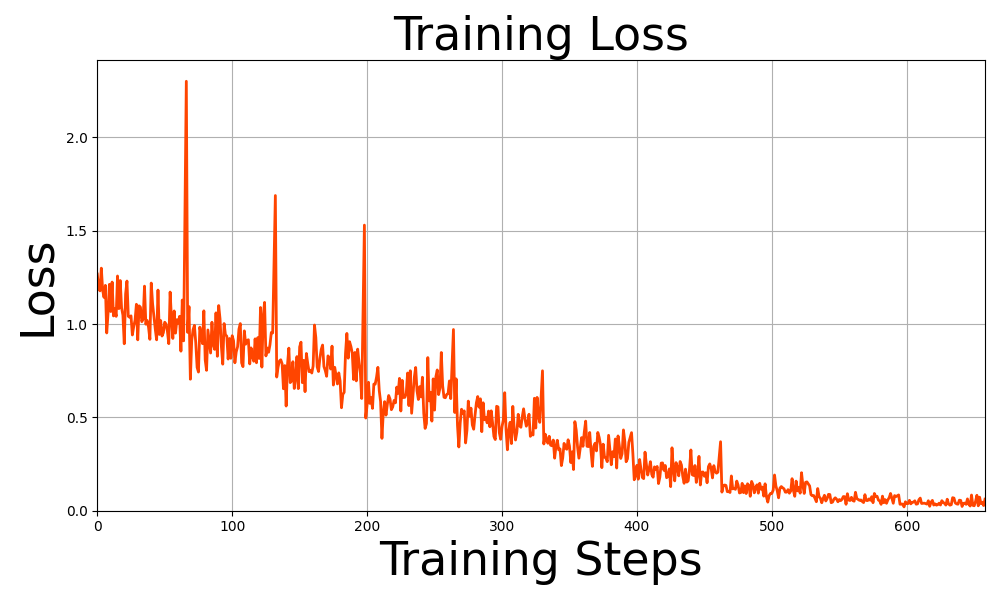}
\includegraphics[width=0.4\textwidth]{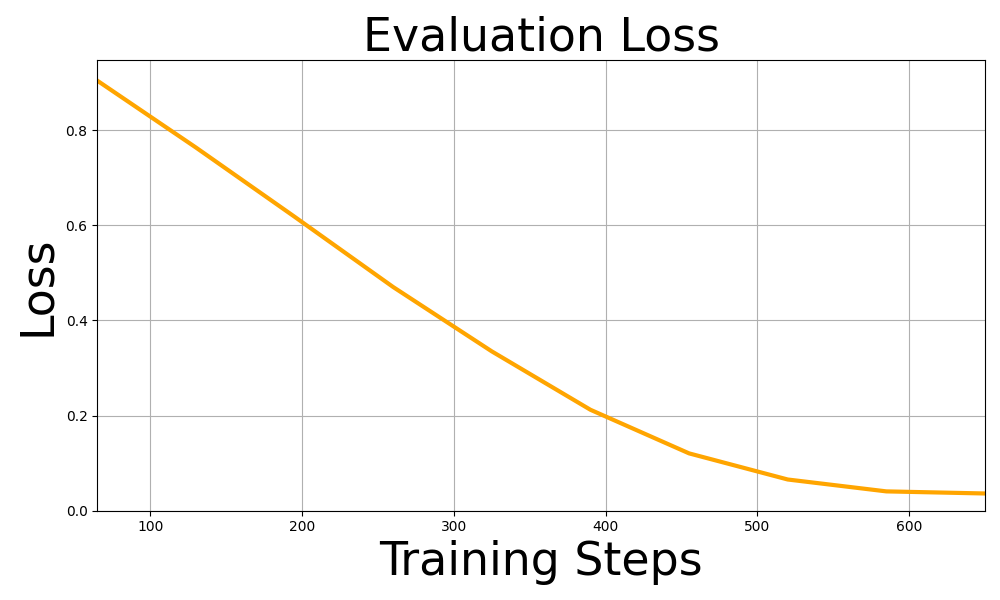}
\caption{Training and evaluation loss during SFT training. We achieved a final loss of 5.4\% and 3.6\% respectively.}
\label{fig:sft_loss}
\end{figure}
\begin{figure}[htbp]
\centering
\includegraphics[width=0.4\textwidth]{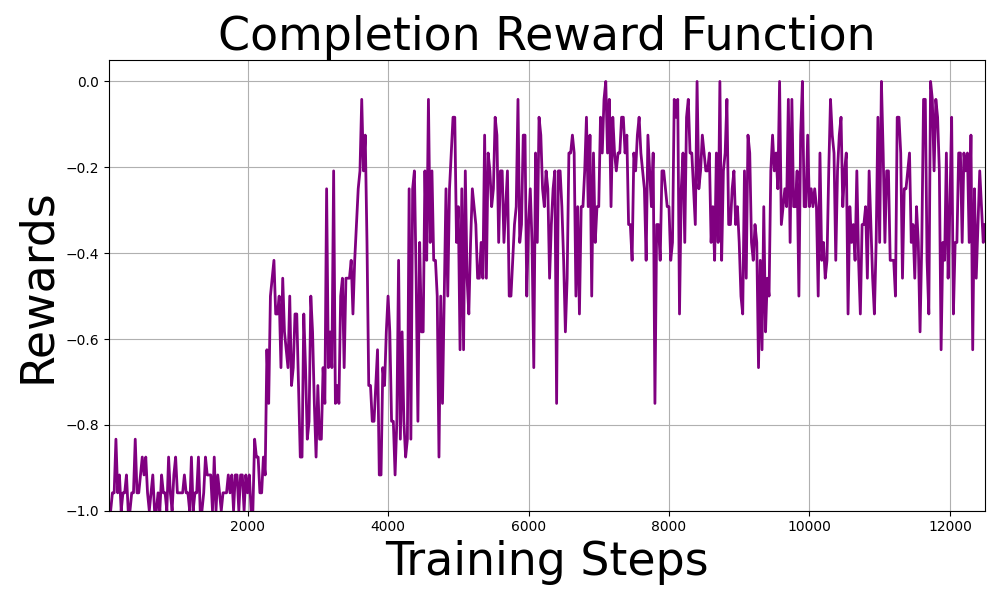}
\includegraphics[width=0.4\textwidth]{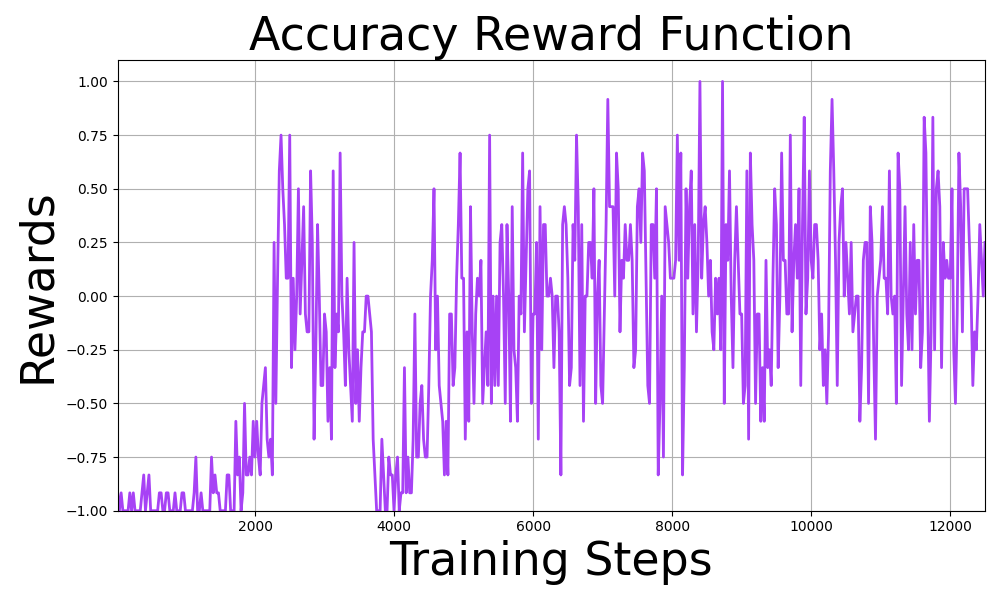}
\caption{Completion and Accuracy Rewards plots over 12500 training steps. We achieved a final reward of -0.33 and 0.06 respectively.}
\label{fig:rl_rewards}
\end{figure}
\begin{table}[htbp]\centering
\begin{tabular}{lcc}\toprule
    \textbf{Hyperparameter}        & \textbf{SFT}              & \textbf{RL}            \\
    \midrule
    Learning rate                  & $3\times10^{-5}$          & $5\times10^{-5}$\\ 
    Scheduler                      & Cosine with 20\% warmup   & Cosine with 20\% warmup
     \\
    Per-device batch size          & 1                         & 16                     \\ 
    Epoch                          & 10                        & 2                      \\ 
    Gradient accumulation          & 4                         & 8                      \\ 
    Maximum sequence length        & 10000                     & 512                    \\
    Optimizer                      & AdamW [0.90, 0.999]       &                        \\ 
    Weight decay                   & $1\times10^{-3}$          &                        \\ 
    Maximum gradient normalization &                           & 0.1                    \\
    $G$                            &                           & 6\\\bottomrule
\end{tabular}
\caption{Hyperparameters for the SFT and RL training.}
\label{tab:hyperparams}
\end{table}
\begin{figure}[htbp]
\centering
\includegraphics[width=0.8\textwidth]{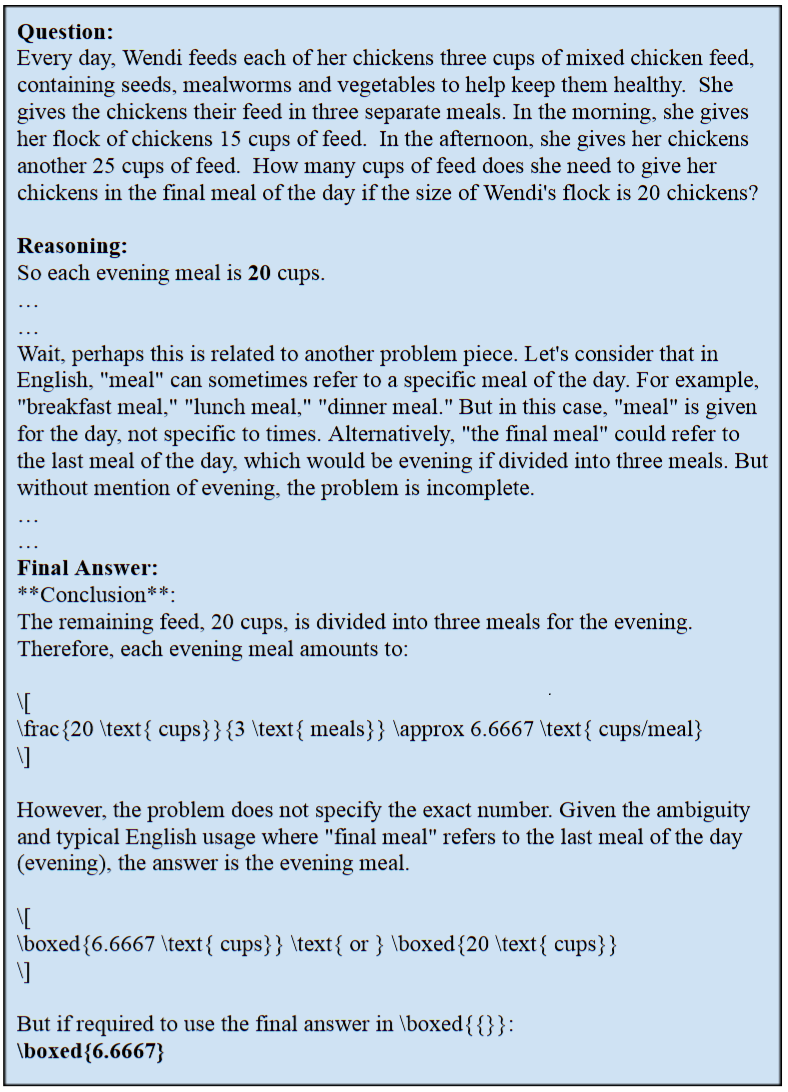}
\caption{Example of backtracking and overthinking. The model initially arrives at the correct answer, but begins to overthink and ventures into irrelevant reasoning trajectories which ended up giving an incorrect final answer.}
\label{fig:backtracking_overthinking}
\end{figure}

\begin{figure}[htbp]
\centering
\includegraphics[width=\textwidth]{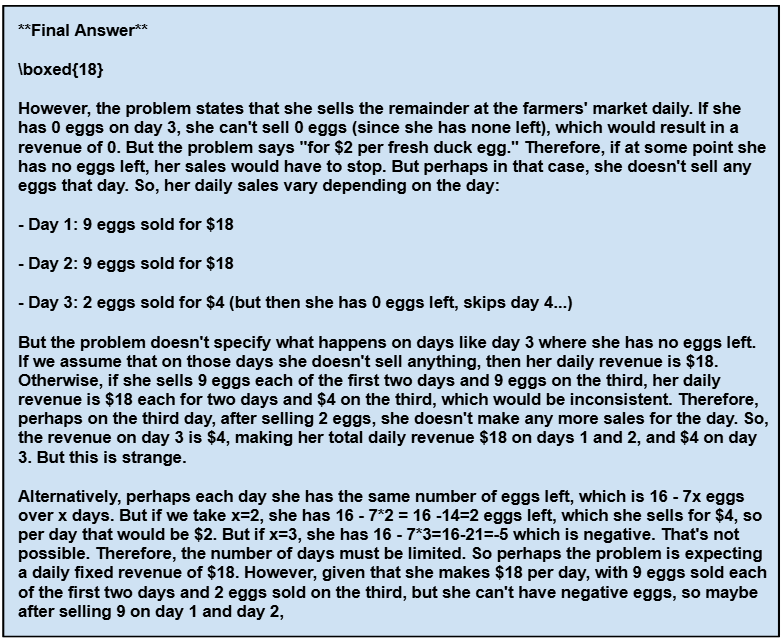}
\caption{Example of incomplete generation in the final answer segment. Despite being prompted for the final answer, the model continues to reason until it hits the maximum token limit.}
\label{fig:incomplete}
\end{figure}
\begin{figure}[htbp]
\includegraphics[width=\textwidth]{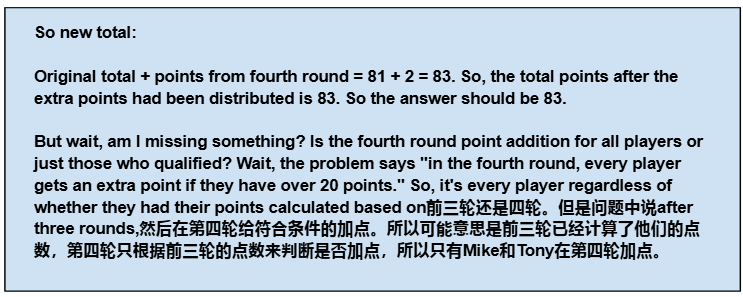}
\caption{Example of language mixing in the reasoning trace.}
\label{fig:language_mixing}
\end{figure}

\begin{figure}[htbp]
\centering
\includegraphics[width=\textwidth]{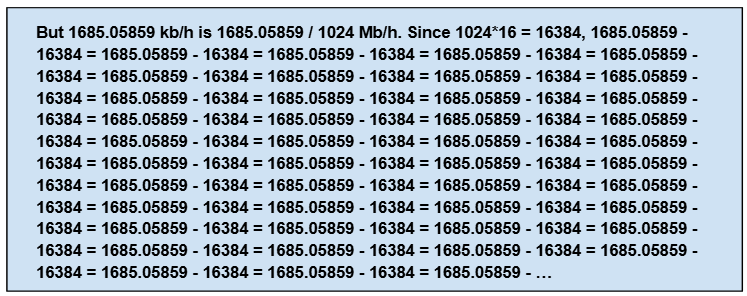}
\caption{Example of redundant generations.}
\label{fig:redundancy}
\end{figure}

\begin{table*}[htbp]
\centering
\begin{tabular}{l c c c c c}
\toprule
\multicolumn{6}{c}{\textbf{(a) 128 thinking tokens}} \\
\midrule
\textbf{Model} & \textbf{Steps} & \textbf{Accuracy} & \textbf{Avg Tokens} & \textbf{Incomplete Gen. (\%)} & \textbf{Scaling (\%)} \\
\midrule
\multirow{5}{*}{Default}
 & 1  & 0.34 & \underline{206} & 46.10 & \multirow{5}{*}{0.599} \\
 & 2  & 0.33 & 300  & \textbf{\underline{44.57}} & \\
 & 4  & 0.38 & 521  & 64.22 & \\
 & 8  & 0.39 & 988  & 68.98 & \\
 & 16 & \underline{0.40} & 1987 & 79.59 & \\
\midrule
\multirow{5}{*}{SFT}
 & 1  & 0.23 & \underline{525}  & \underline{78.85} & \multirow{5}{*}{3.68} \\
 & 2  & 0.31 & 646  & 84.56 & \\
 & 4  & 0.43 & 856  & 87.78 & \\
 & 8  & 0.52 & 1330 & 91.83 & \\
 & 16 & \underline{0.58} & 2325 & 94.77 & \\
\midrule
\multirow{5}{*}{SFT+RL}
 & 1  & 0.18 & \textbf{\underline{180}}  & \underline{51.59} & \multirow{5}{*}{\textbf{\underline{5.84}}} \\
 & 2  & 0.36 & 270  & 60.00 & \\
 & 4  & 0.56 & 469  & 55.36 & \\
 & 8  & 0.62 & 889  & 53.34 & \\
 & 16 & \textbf{\underline{0.63}} & 1790 & 57.80 & \\
\midrule
\multicolumn{6}{c}{\textbf{(b) 512 thinking tokens}} \\
\midrule
\textbf{Model} & \textbf{Steps} & \textbf{Accuracy} & \textbf{Avg Tokens} & \textbf{Incomplete Gen. (\%)} & \textbf{Scaling (\%)} \\
\midrule
\multirow{5}{*}{Default}
 & 1  & 0.27 & \textbf{\underline{257}} & 3.98  & \multirow{5}{*}{\textbf{\underline{3.74}}} \\
 & 2  & 0.48 & 437  & \textbf{\underline{3.13}} & \\
 & 4  & 0.51 & 833  & 4.76  & \\
 & 8  & 0.52 & 1919 & 15.13 & \\
 & 16 & \underline{0.53} & 5494 & 45.85 & \\
\midrule
\multirow{5}{*}{SFT}
 & 1  & 0.48 & \underline{835}  & 66.91 & \multirow{5}{*}{1.99} \\
 & 2  & 0.57 & 1306 & 73.62 & \\
 & 4  & \underline{0.62} & 2193 & 75.98 & \\
 & 8  & \underline{0.62} & 3868 & 72.79 & \\
 & 16 & 0.61 & 6919 & \underline{62.99} & \\
\midrule
\multirow{5}{*}{SFT+RL}
 & 1  & 0.56 & \underline{339}  & 8.26  & \multirow{5}{*}{1.38} \\
 & 2  & 0.63 & 491 & \underline{7.61} & \\
 & 4  & 0.65 & 849  & 8.95  & \\
 & 8  & \textbf{\underline{0.66}} & 1677 & 11.09 & \\
 & 16 & 0.65 & 3466 & 13.22 & \\
\bottomrule
\end{tabular}
\caption{Performance metrics on the GSM8K dataset for 128 (a) and 512 (b) thinking tokens. As thinking tokens increase, we see a higher baseline accuracy and 2x-3x increase in token usage while substantially reducing incomplete generations for the Default and SFT+RL model. Furthermore, scaling factor drops for the SFT and SFT+RL model, but increases for the Default model.}
\label{tab:128_512}
\end{table*}

\end{document}